\documentclass{elsarticle}

\usepackage{graphicx}
\usepackage{listings}
\usepackage{booktabs}
\usepackage{multirow}
\usepackage{tabularx}
\usepackage{xcolor}
\usepackage{amssymb}
\usepackage{amsmath}

\usepackage{natbib}

\setcitestyle{comma}

\begin{document}

\begin{frontmatter}

\title{Learning to Select Goals in \\Automated Planning with Deep-Q Learning}

\author{Carlos Núñez-Molina}
\ead{ccaarlos@ugr.es}
\author{Juan Fernández-Olivares}
\ead{faro@decsai.ugr.es}
\author{Raúl Pérez}
\ead{fgr@decsai.ugr.es}

\address{Universidad de Granada, Spain}

\begin{abstract}

{
In this work we propose a planning and acting architecture endowed with a module which learns to select subgoals with Deep Q-Learning. This allows us to decrease the load of a planner when faced with scenarios with real-time restrictions.
We have trained this architecture on a video game environment used as a standard test-bed for intelligent systems applications, testing it on different levels of the same game to evaluate its generalization abilities. We have measured the performance of our approach as more training data is made available, {as well as} compared it with both a state-of-the-art, classical planner and the standard Deep Q-Learning algorithm.
The results obtained show our model performs better than the alternative methods considered, when both plan quality (plan length) and time requirements are taken into account. On the one hand, it is more sample-efficient than standard Deep Q-Learning, and it is able to generalize better across levels. On the other hand, it reduces problem-solving time when compared with a state-of-the-art automated planner, at the expense of obtaining plans with only 9\% more actions.
}

\end{abstract}
	
\begin{keyword}
{
automated planning \sep goal selection \sep deep Q-learning
}
\end{keyword}
	
\end{frontmatter}

\section{Introduction}
\label{section:introduction}

\begingroup
\renewcommand{\thefootnote}{\relax}
\footnotetext{This work has been published in the journal \textit{Expert Systems with Applications}: doi.org/10.1016/j.eswa.2022.117265}
\endgroup

Automated Planning (AP) \citep{ghallab2016automated} is a subfield of Artificial Intelligence devoted to providing goal-oriented, deliberative behaviour to both physical and virtual agents, e.g. robots or video game automated players. An automated planner takes as input a planning domain, an initial state and a goal and carries out a search process that returns a plan (sequence  of actions) that guides the behaviour of the agent, in order to reach the given goal from the initial state. The planning domain describes the actions an agent can execute, as well as the dynamics of the environment where the agent is expected to act. The initial state is a set of facts describing the context in which the agent initiates its behaviour, and the goal is a set of conditions which need to be accomplished at the end of the plan.

Automated Planning has traditionally been one of the most widely used techniques in AI and has been successfully applied in real-world applications  \citep{castillo2008samap, fdez2019personalized}. {However, in order to integrate it
into online execution systems, i.e., systems used in real-time
scenarios which interleave planning and acting, there exist
several issues which must be addressed. The most important one is that planning is often too slow for real-time scenarios. In most real-world
problems the search space grows exponentially with problem size so, despite the use
of heuristics, finding a suitable plan usually takes very long.}

Due to this, despite great advances in the integration of planning and acting into online architectures \citep{patra2019acting}, most recent works which apply AI to guide the behaviour of agents in real-time scenarios, like video games, do not integrate planning into their agent architecture. This can be clearly seen in \citep{vinyals2019grandmaster}. In this impactful work, an agent is trained to play \emph{Starcraft}, a highly competitive real-time strategy (RTS) game. This seems like a perfect problem for planning: players need to establish a long-term, goal-oriented strategy in order to achieve victory and all the dynamics of the game are known, so they can be represented into a planning domain. {However, the authors decided to integrate Deep Learning \citep{lecun2015deep} with Reinforcement Learning \citep{sutton2018reinforcement} to model the behaviour of the agent, instead of recurring to Automated Planning.}

Architectures which rely on Machine Learning (ML) and Reinforcement Learning (RL) present some advantages over planning: they usually require very little prior knowledge about the domain (they do not need a planning domain) and, once trained, they act quickly, since they do not perform any type of planning. Nevertheless, they also have some drawbacks. Firstly, they are very sample inefficient. They require a lot of data in order to learn, in the order of hundreds of thousands or even millions samples \citep{torrado2018deep}. Secondly, they usually present bad generalization properties, i.e., have difficulties in applying what they have learned not only to new domains but also to new problems of the same domain \citep{zhang2018study}.

Since both Automated Planning and Reinforcement Learning have their own pros and cons, it seems natural to try to combine them as part of the same agent architecture, which ideally would possess the best of both worlds. {Aligned with that purpose, the main contribution of this work is the proposal of a Goal Selection Module based on Deep Q-Learning, and its integration into a planning and acting architecture to control the behaviour of an agent, in a real-time
environment. We have tested our approach on the {General Video Game AI (GVGAI) framework \citep{perez20152014}}, measuring its generalization abilities across different levels of the same game.}

{
The results of our experiments show how our approach is able to exploit the synergy between Automated Planning and Reinforcement Learning, improving the performance of each of these techniques when applied separately. On the one hand, by using subgoals as the units of action to choose in RL, we are able to improve sample efficiency and generalization. Our approach obtains plans of better quality (plan length) than standard Deep Q-Learning, while being trained on a dataset ten times smaller. On the other hand, it also reduces problem-solving time when compared with a state-of-the-art automated planner, at the expense of obtaining plans with only 9\% more actions. {Moreover, the planning and acting architecture, endowed with our Goal Selection module, is able to solve every test level of the game used in this work in less than 2 seconds per level, even those which the same planner alone (without our goal selection module) cannot solve in 1 hour.}
}

The structure of this work is the following. {In Section \ref{section:background}, we
explain the background concepts needed to understand the remainder of the paper. Then, we present an overview of our agent architecture {in Section \ref{section:the_planning_and_acting_architecture},} and explain how it learns to select subgoals in Section \ref{section:goal_selection_learning}. After this, we present the results of our experiments in Section \ref{section:experiments_and_analysis_of_results}. In Section \ref{section:related_work}, we compare our approach with related work. Finally, in Section \ref{section:conclusions_and_future_work}, we present our conclusions and outline possible future work.
}

\section{Background}
\label{section:background}
{
This section briefly describes some background concepts needed to understand our work. We firstly explain the GVGAI video game framework and the game used in this work. Then, we describe PDDL, the standard planning language used to represent action knowledge and which our architecture uses. Finally, we outline some basic concepts about Reinforcement Learning and 
{Deep Q-Learning, needed to understand the technical contributions of Section \ref{section:goal_selection_learning}.}
}

\subsection{GVGAI and Boulder Dash}
\label{subsection:gvgai}

To test our planning and acting architecture we have used the General Video Game AI (GVGAI) Framework \citep{perez20152014}. This framework provides a game environment with a large quantity of tile-based games which are also very different in kind. For example, it comprises purely reactive games, such as \emph{Space Invaders}, and also games which require deliberative, long-term planning in order to be solved successfully, like \emph{Sokoban}. {Additionally, levels of GVGAI games are represented as plain text files (known as level description files), which allows us to create as many levels as we need to train and test our approach.}

\begin{figure}[h]
	\centering
	\includegraphics[width=\linewidth]{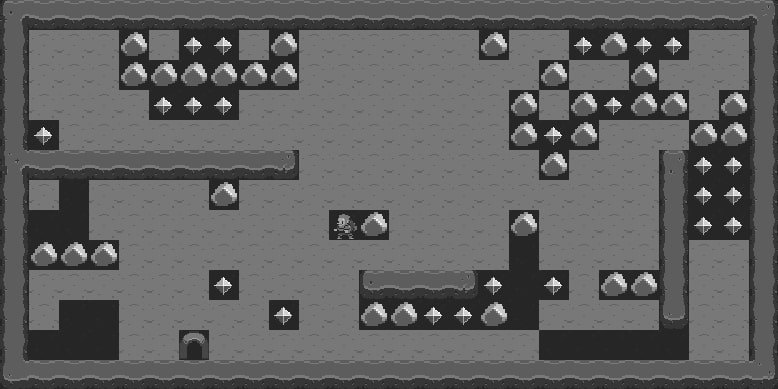}
	\caption{\textbf{A level of the Boulder Dash game.}}
	\label{fig:boulder_dash}
\end{figure}

In this work, we have decided to use a deterministic version of the GVGAI game known as \emph{Boulder Dash} (\emph{see Figure \ref{fig:boulder_dash}}). We use this game to extract the experience of episodes of planning and acting our Goal Selection Module is trained on.
{
In our version of this game, there are no enemies and boulders do not fall. The goal of the player is to collect nine gems and then get to the exit, while minimizing the number of actions used.} 

{
There are five different actions available to the player: \emph{UP}, \emph{DOWN}, \emph{LEFT}, \emph{RIGHT}, and \emph{USE}. The four first actions let the player traverse the level, one tile at a time. The last action, \emph{USE}, is used by the player to break a boulder with its pickaxe before passing through. The player can be oriented in four different directions: \emph{NORTH}, \emph{SOUTH}, \emph{EAST} or \emph{WEST}. When the player uses a movement action, it turns towards the corresponding direction or, if it was already facing that way, it moves one tile. For instance, if the player executes action \emph{UP} and was facing \emph{SOUTH}, it will now face \emph{NORTH}. But, if it was already facing \emph{NORTH}, it will move up one tile.
}

\lstset{
    caption={\textbf{Level description file of the level shown in Figure \ref{fig:boulder_dash}.} Each letter represents a different type of object: ``w'' for walls, ``o'' for boulders, ``x'' for gems, ``A'' for the player, ``e'' for the exit and ``.'' for dirt. The character ``-'' represents empty tiles.},
    captionpos=b,
    xleftmargin=.13\textwidth, xrightmargin=.13\textwidth
}
\begin{lstlisting}
  
	wwwwwwwwwwwwwwwwwwwwwwwwww
	w...o.xx.o......o..xoxx..w
	w...oooooo........o..o...w
	w....xxx.........o.oxoo.ow
	wx...............oxo...oow
	wwwwwwwwww........o...wxxw
	w.-....o..............wxxw
	w--........Ao....o....wxxw
	wooo.............-....w..w
	w......x....wwwwx-x.oow..w
	w.--.....x..ooxxo-....w..w
	w---..e...........-----..w
	wwwwwwwwwwwwwwwwwwwwwwwwww
\end{lstlisting}

{
{We have used a static version of Boulder Dash because we need a controllable environment to conduct the experimentation of our proposal.} This way, we can test and validate our goal selection method in an isolated way, without having to deal with dynamism or uncertainty. For instance, in the original version of Boulder Dash boulders may fall and kill the agent. If this were also the case for our version of the game, we could not assume that a valid plan is always successful, i.e., always takes the agent from the current state of the game to a state where the corresponding subgoal has been achieved. Thus, our architecture would additionally need to detect and manage risks associated with the execution of plans in environments with uncertainty, which is outside the scope of this work. Nevertheless, in Section \ref{section:conclusions_and_future_work} we propose as future work a method for managing uncertainty by integrating Automated Planning with Deep Q-Learning. This method would allow our approach to be applied in dynamic and uncertain environments, including the original version of Boulder Dash.
}

{
Furthermore, our version of Boulder Dash still represents a great challenge for RL and AP techniques, as the results of Section \ref{subsection:discussion} show. Every level contains 23 gems, but the agent only needs to obtain 9 of them. If we assume the agent always obtains 9 gems and then goes to the exit, then the number of total possible trajectories is given by the following expression\footnote{The set of 9 subgoals can be achieved in $9!$ different ways by the agent.}: $ \binom{23}{9}*9! = 296.541.907.200$. Therefore, there are more than 200 billion different trajectories for a single Boulder Dash level. This combinatorial explosion means Boulder Dash is very hard to solve, even if it contains no uncertainty at all.
}

\subsection{PDDL}
\label{subsection:pddl}

{We have used PDDL (Fox and Long, 2003) to encode the inputs for our Planner Module. This is a standard language used in Automated Planning for representing planning domains and problems.} The PDDL Domain file contains the information about the planning domain: a description of the predicates used to represent a problem state and the preconditions and effects of every action. The PDDL Problem file contains the description of a given planning problem: a representation of its initial state and the goal to achieve. A planner receives as inputs these two files and returns a plan, which is constituted by an ordered sequence of instantiated PDDL actions.

Given a GVGAI game, we can create its associated planning domain. This domain will encode the game dynamics, i.e., the different entities of the game (Listing 2) and all the actions the player can do to interact with them (Listing 3). Each game level will have a different planning problem associated, representing its initial state (Listing 4) and the goal to achieve (Listing 5). For instance, in Boulder Dash a goal corresponds to getting a gem present at the level or getting to the exit.

  \lstset{
    caption=\textbf{Predicates used to represent the position of entities in the Boulder Dash domain.} {The predicate \textit{at} represents which cell an object is in. The predicate \textit{connected-*} encodes the adjacent cells to a given cell.},
    captionpos=b
  }
  \begin{lstlisting}
(:predicates
	(at ?l - Locatable ?c - Cell)
	(connected-up ?c1 ?c2 - Cell)
	(connected-down ?c1 ?c2 - Cell)
	(connected-left ?c1 ?c2 - Cell)
	(connected-right ?c1 ?c2 - Cell)
)
  \end{lstlisting}
  
    \lstset{
    caption=\textbf{Preconditions and effects of the \emph{move-up} action in the Boulder Dash domain.} This action moves the player  one cell \emph{up} (north). The player must be oriented \emph{up} and there cannot be a wall or a boulder in the cell it is going to.,
    captionpos=b
  }
  \begin{lstlisting}
(:action move-up
	:parameters (?p - Player ?c1 ?c2 - Cell)
	:precondition (and
	  (at ?p ?c1)
	  (oriented-up ?p)
	  (connected-up ?c1 ?c2)
	  (not (exists (?b - Boulder) 
	       (at ?b ?c2)))
	  (not (terrain-wall ?c2))
	)
	:effect (and
	  (when
	    (not (terrain-empty ?c2))
	    (terrain-empty ?c2)
	  )
	  (not (at ?p ?c1))
	  (at ?p ?c2)
	)
)
  \end{lstlisting}

  \lstset{
    caption=\textbf{Part of the predicates describing the initial state of a given Boulder Dash problem.} The gem \emph{gem1} is in cell \emph{c\_5\_3}{,} which is next to cells \emph{c\_5\_2}{,} \emph{c\_5\_4}{,} \emph{c\_6\_3} and \emph{c\_4\_3}.,
    captionpos=b
  }
  \begin{lstlisting}
(:init
	(at gem1 c_5_3)
	(connected-up c_5_3 c_5_2)
	(connected-down c_5_3 c_5_4)
	(connected-right c_5_3 c_6_3)
	(connected-left c_5_3 c_4_3)
)
  \end{lstlisting}
  
  \lstset{
    caption=\textbf{Goal description of a given Boulder Dash problem.} In this problem{,} the player must pick the gem \emph{gem13} (it must get to the cell where \emph{gem13} is in).,
    captionpos=b
  }
  \begin{lstlisting}
(:goal
	  (got gem13)
)
  \end{lstlisting}

\subsection{Deep Q-Learning}
\label{subsection:deep-q-learning}

{
The version of Boulder Dash used in this work can be described by a Deterministic Markov Decision Process (DMDP), which is a special case of MDP \citep{heuristic2012} where executing an action $a$ in a state $s$ always leads into the same next state $s'$. A DMDP is represented by a 4-tuple $(S, A, r, t)$, where $S$ is the state space, $A$ is the action space or set of actions the agent can execute, $r: S \times A \rightarrow \mathbb{R}$ is the immediate reward function, and $t: S \times A \rightarrow S$
is the transition function, which determines the next state $s'$ of the environment when action $a$ is executed in state $s$. A deterministic policy $\pi : S \rightarrow A$ solves a DMDP by 
mapping each state $s \in S$ to an action $a \in A$.  {The cumulative reward $R$ is the sum of the immediate rewards $r(s, a)$, obtained by the agent when it selects in each state $s$ the action $\pi(s) = a$, given by its
policy until the end of the episode.} The optimal policy $\pi^*$ is the policy which maximizes the cumulative reward $R$.
}

{
Reinforcement Learning (RL) \citep{sutton2018reinforcement} comprises 
a family of algorithms which can be used to learn this optimal policy $\pi^*$ from data. One of the most widely used RL techniques is known as Q-Learning \citep{watkins1989learning}. It learns a value for each $(s,a)$ pair, known as the Q-value $Q(s,a)$, which represents how good action $a$ is when executed in state $s$. Thus, the optimal policy $\pi^*$ consists in selecting for each state the action with the highest Q-value. One of the main problems Q-Learning has is that it needs to learn the Q-value $Q(s,a)$ for each possible $(s,a)$ pair, which together constitute the Q-table. If the action or state space are too large, the Q-table grows and the learning problem becomes intractable. Deep Q-Learning (DQL) \citep{mnih2013playing} solves this problem by making use of a Deep Neural Network which learns the Q-values. This way, DQL is able to generalize and correctly predict the Q-values for new $(s,a)$ pairs never seen before by the network. This is why we use DQL instead of traditional Q-Learning in our work, in pursuit of these generalization abilities.
}

{
Given the current state $s$, DQL predicts a value $Q(s,a)$ for each action $a \in A$ and selects the action $\hat{a}$ with the highest Q-value. {The Q-value $Q(s, a)$ represents the immediate reward $r(s, a)$ plus the cumulative reward $R$, obtained by following the optimal policy $\pi^*$
from the next state $s'$ until the end of the
episode.} Since its correct value, known as the Q-target $Q^*(s,a)$, is unknown, we need to utilize the Bellman Equation, which recursively approximates the Q-targets $Q^*(s,a)$ using other Q-values $Q(s',a')$. The loss function $L$ used by Deep Q-Learning incorporates this to define the error to be minimized:
}

{
\begin{equation}
\label{equation:deep-q-learning}
L = (Q(s,a) - Q^*(s,a))^2 = (Q(s,a) - (r + \gamma \max_{a' \in A'} Q(s',a')))^2
\end{equation}
}

\noindent {where $s$ is the current state, $a \in A$ an action which can be executed in $s$, $r$ the immediate reward obtained by applying $a$ in $s$, $s'$ the next state, and $\gamma \in [0,1]$ a discount factor which determines the relative importance of immediate rewards vs future rewards.}

\section{The Planning and Acting Architecture}
\label{section:the_planning_and_acting_architecture}

\begin{figure}[h]
	\centering
	\includegraphics[width=.7\linewidth]{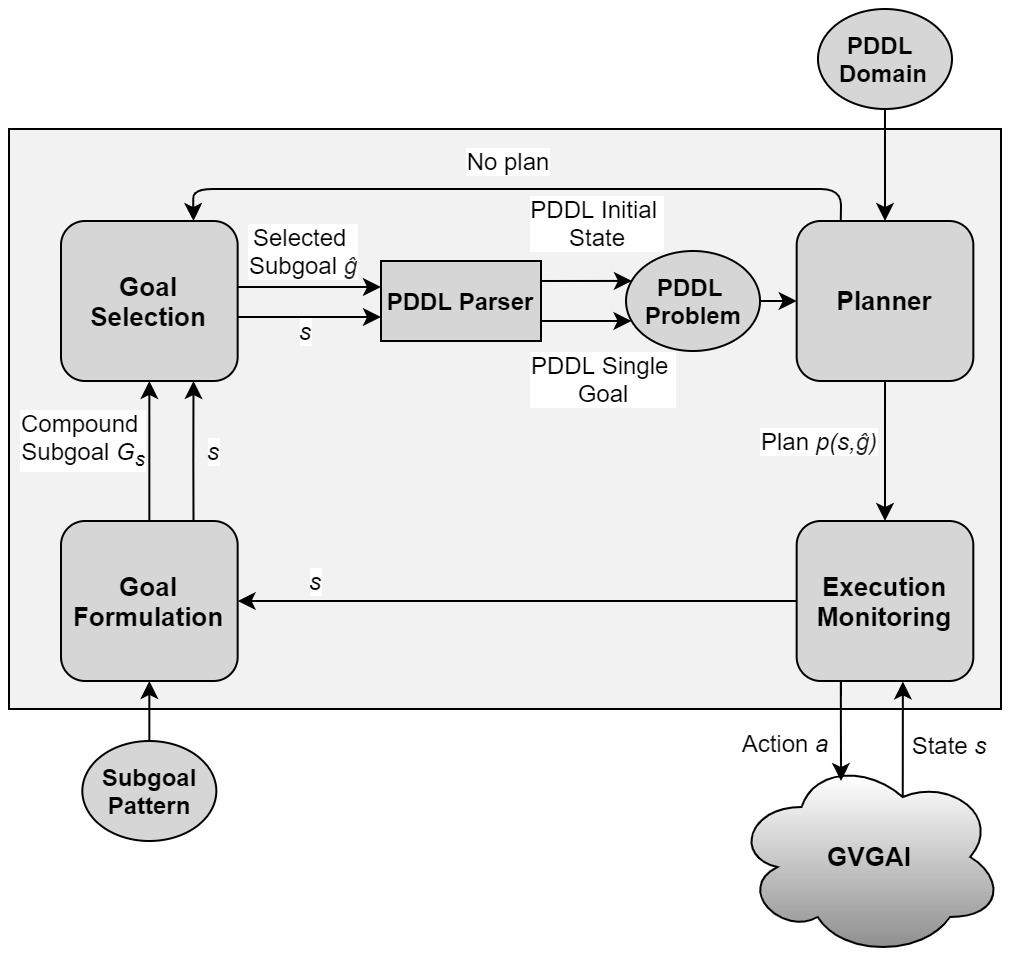}
	\caption{\textbf{An overview of the planning and acting architecture.}}
	\label{fig:architecture}
\end{figure}

An overview of the planning and acting architecture can be seen in Figure \ref{fig:architecture}. The \textbf{Execution Monitoring} Module communicates with the GVGAI environment, receiving the current state $s$ of the game. It also supervises the state of the current plan. If it is not empty, it returns the next action $a$. If it is empty, the architecture needs to find a new plan. The \textbf{Goal Formulation} Module
receives $s$ and generates the compound subgoal $G_s$, which is the set of the eligible subgoals $g_1, g_2, ..., g_n$ that can be selected in the state $s$. The final goal $g_f$ is also included in $G_s$.
The \textbf{Subgoal Pattern} contains the prior information about the domain needed to automatically generate $G_s$ given $s$. In Boulder Dash each subgoal $g$ corresponds to getting one of the available gems in $s$, and the final goal $g_f$ corresponds to getting to the exit.
Since all GVGAI games are tile-based, we have associated each subgoal {(and also the final goal)} with getting to its correspondent tile (cell). The \textbf{Goal Selection} Module receives the compound subgoal $G_s$, and selects the best subgoal $\hat{g} \in G_s$ given $s$. The \textbf{PDDL Parser} encodes $\hat{g}$ as a PDDL Single Goal, i.e., (\emph{got gem13}), and $s$ as a PDDL Initial State, which together make up the PDDL Problem (\emph{see example in Section \ref{subsection:pddl}}). The \textbf{Planner} Module receives the PDDL Problem along with the PDDL Domain, provided by a human expert, and generates a plan $p(s,\hat{g})$ which achieves $\hat{g}$ starting from $s$.
{In case the Goal Selection Module has selected as $\hat{g}$ a subgoal which cannot be reached from the state $s$, the Planner Module will not be able to find a valid plan $p(s,\hat{g})$, and will return an empty one. We will call this a \emph{goal selection error}. In Boulder Dash, a goal selection error happens when the Goal Selection Module chooses the final goal, i.e., $\hat{g} = g_f$, but the agent has not obtained nine gems yet. When this happens, the Goal Selection Module must select a new subgoal $\hat{g}$. Once a valid plan $p(s,\hat{g})$ has been found by the planner}, the Execution Monitoring Module receives it and the cycle completes.

\section{Goal Selection Learning}
\label{section:goal_selection_learning}
{
This section {firstly} describes the underlying model of the goal selection problem, addressed as a sequential decision making problem, and based on a DMDP (\emph{see Section \ref{subsection:deep-q-learning}}). {Then, a detailed description of the CNN architecture, used by the Goal Selection Module, and the training process are shown.}
}

\subsection{Formulation of Goal Selection as a Deterministic MDP}
\label{subsection:math_formulation}

{In this work, the goal of the planning and acting architecture is to achieve the final goal $g_f$, i.e., complete a game level, using the minimum possible number of actions. To achieve this, it must select a series of subgoals $g_1, g_2, ..., g_n, g_f$ in the optimal order to complete the level using as few actions as possible. In order to select the best subgoal $\hat{g}$ for the current game state $s$, the Goal Selection Module iterates over every eligible subgoal $g \in G_s$, and predicts the length $l_{P(s,g)}$ of the \emph{total plan} associated with each one. This value $l_{P(s,g)}$ corresponds to the length of the plan $P(s,g)$ which, starting from $s$, achieves $g$ and, once obtained it, then achieves the final goal $g_f$ (after obtaining the required subgoals in an optimal order). The best subgoal $\hat{g}$ corresponds to the subgoal for which has been predicted the minimum length $l_{P(s,\hat{g})}$. This is the one selected by the Goal Selection Module.}

{
To predict this $l_{P(s,g)}$ value for a given $(s,g)$ pair, the Goal Selection Module integrates a Convolutional Neural Network (CNN) \citep{krizhevsky2012imagenet}. The Neural Network receives as input a state $s$ of the game and a subgoal $g \in G_s$, both encoded as a three-dimensional tensor, which will be referred to as the \emph{one-hot tensor}, and outputs the number of actions of the associated total plan $P(s,g)$. The first two dimensions of this one-hot tensor are associated with the $(x,y)$ position of a game tile. The third one is used to encode the object present at that game tile. The information about the objects is encoded as a one-hot vector. In our version of Boulder Dash, there are six different types of objects: \emph{player}, \emph{exit}, \emph{boulder}, \emph{gem}, \emph{wall} and \emph{dirt}. Each different type is associated with a number $i \in \{1..6\}$, representing the \emph{i-th} position of the one-hot vector. If an object of type $i$ is present at the $(x,y)$ tile of the level, then the one-hot tensor position $(x,y,i)$ contains a value of 1. To represent subgoals $g$, we simply treat them as an additional type of object with corresponding number $i=7$. This way, if the subgoal $g$ of the one-hot tensor that represents the pair $(s,g)$ is at the $(x,y)$ tile of the level, then the one-hot tensor will contain a value of 1 in its $(x,y,7)$ position.
}

{
Unlike most RL problems where the action space is fixed, i.e., the agent always has the same set of actions to choose from, in our problem the number of eligible subgoals depends on the current state $s$ of the game. In addition, each level will contain a different initial set of subgoals, i.e., the subgoals will be in different positions. For this reason, the CNN needs to receive both $s$ and $g$ (encoded as the one-hot tensor $(s,g)$) as inputs, and {learns} to reason about states and subgoals, being capable of generalizing to both new states and subgoals.
}

{
We can associate two different DMDPs to our version of Boulder Dash: 
$M = (S, A, r, t)$ and $M^g = (S^g, G_s, r^g, t^g)$. {$M$ corresponds to the standard RL description detailed in Section \ref{subsection:deep-q-learning}. $M^g$ is a special type of DMDP which describes the game from the perspective of selecting subgoals $\hat{g} \in G_s$ instead of executing actions $\hat{a} \in A$.} This alternative formulation $M^g$ is the one our planning and acting architecture is built upon. The correspondence between $M$ and $M^g$ is detailed below:
}

{
\begin{itemize}
    \item $S^g$ is the state space of $M^g$, which only contains a subset of the states present in the state space $S$ of $M$, i.e., $S^g \subset S$. $S^g$ contains the initial state of $S$ and also the final states $s'$ of the plans $p(s,g)$ which achieve the subgoals, where $s \in S^g$ and $g \in G_s$.
    \item $G_s$ is the compound subgoal, i.e., the set of eligible subgoals the agent can choose from in state $s$. It always contains the final goal $g_f$. $G^{\circ}_s \subset G_s$ is the set of attainable subgoals for state $s$, containing the subgoals and/or final goal $g$ for which there exist a valid plan $p(s,g)$ that achieves $g$ starting from state $s$. If the subgoal selected by the agent is not attainable, the Planner Module will not be able to find a valid plan and a goal selection error will be produced (\emph{see Section \ref{section:the_planning_and_acting_architecture}}).
    \item $r^g : S \times G_s \rightarrow \mathbb{R}$ is the immediate reward, which depends on the state $s$ and the selected subgoal $\hat{g} \in G_s$. If the subgoal is attainable, i.e., $\hat{g} \in G^{\circ}_s$, then the immediate reward is equal to the length $l_{p(s,\hat{g})}$ of the plan $p(s,\hat{g})$ which achieves $\hat{g}$ starting from $s$. If $\hat{g}$ is not attainable, then the reward is equal to a value $\lambda$ which serves as a penalization for the agent.
    \item $t^g: S \times G^{\circ}_s \rightarrow S$ is the transition function, which determines the next state $s'$ of the environment when the agent selects an attainable subgoal $\hat{g} \in G^{\circ}_s$ in state $s$. The next state $s'$ corresponds to the final state of the plan $p(s, \hat{g})$ which achieves $\hat{g}$ starting from $s$. Since the planner used to obtain $p(s, \hat{g})$ is deterministic, i.e., for a given $(s, \hat{g})$ it always obtains the same plan, this means that the transition function $t^g$ is also deterministic.
\end{itemize}}

{
After defining all the elements of $M^g$ we can now adapt the cumulative reward $R$ and the deterministic policy $\pi$ to this special type of DMDP. A deterministic policy $\pi^g : S^g \rightarrow G_s$ maps each state $s \in S^g$ to an eligible subgoal $g \in G_s$. The cumulative reward $R^g$ is the sum of the immediate rewards $r^g(s,g)$ obtained by the agent when it selects in each state $s$ the subgoal $\pi^g(s) = \hat{g}$, given by its policy, until the end of the episode. 
{
The optimal policy $\pi^{g^*}$ is the one which minimizes $R^g$, and represents the optimal sequence of subgoal selections, where the last subgoal is always the final goal $g_f$. The reason for minimizing $R^g$, instead of maximizing it, comes from the fact that the cumulative reward $R^g$ represents the sum of the lengths of the plans $p(s,\hat{g})$ (in case all selected subgoals are attainable). Since our goal is to solve the levels using the minimum possible number of actions, we are interested on minimizing this quantity. Finally, it is worth mentioning that one of the main advantages of using $M^g$ to formulate and solve a RL problem, instead of a standard DMDP description, is that the state space is reduced (since $S^g \subset S$) and, thus, the problem is simplified.} The implications of this will be explored in Section \ref{subsection:discussion}.}

{
Using the formulation given by $M^g$, we can adapt the DQL algorithm to this new type of DMDP. In this case, DQL predicts a Q-value $Q(s,g)$ for each $(s,g)$ pair, where $s \in S^g$ and $g \in G_s$, and selects the subgoal $\hat{g}$ with the lowest Q-value. This Q-value $Q(s,g)$ represents the immediate reward $r^g(s,g)$ plus the cumulative reward $R^g$, obtained by following the optimal policy $\pi^{g^*}$, from the next state $s'$ until the end of the episode.
{If the subgoal is attainable ($g \in G^{\circ}_s$), then $Q^*(s,g) = l_{P(s,g)}$. That is to say, the correct Q-value, i.e., the Q-target $Q^*(s,g)$, is equal to the length of the total plan $P(s,g)$ associated with the $(s,g)$ pair. If $g$ is not attainable, then $Q^*(s,g) = r^g(s,g) = \lambda$, i.e., the Q-target is equal to the penalization $\lambda$.}
As happens with standard DQL, the value of the Q-target is unknown, and needs to be recursively estimated using the Bellman Equation. Thus, the loss function $L^g$ associated with the DMDP $M^g$ is as follows:
}

{
\begin{equation}
\label{equation:deep-q-planning}
L^g = \Big(Q(s,g) - Q^*(s,g)\Big)^2 = 
\begin{cases}
\Big(Q(s,g) - l_{P(s,g)}\Big)^2 \text{, \qquad if $g \in G^{\circ}_s$.}\\ \\
\Big(Q(s,g) - \lambda\Big)^2 \text{, \qquad if $g \notin G^{\circ}_s.$}
\end{cases}
\end{equation}
}

\noindent {where $\Big(Q(s,g) - l_{P(s,g)}\Big)^2 = \Big(Q(s,g) - \big(l_{p(s,g)} + \gamma \underset{g' \in G_{s'}}{\min} Q(s',g')\big) \Big)^2$, $s$ is the current state, $g \in G_s$ is an eligible subgoal in state $s$, $s'$ is the next state and $\gamma \in [0,1]$ is the discount factor.}

\subsection{CNN Architecture and Training}
\label{subsection:cnn_architecture}

\begin{figure}[h]
	\centering
	\includegraphics[width=\linewidth]{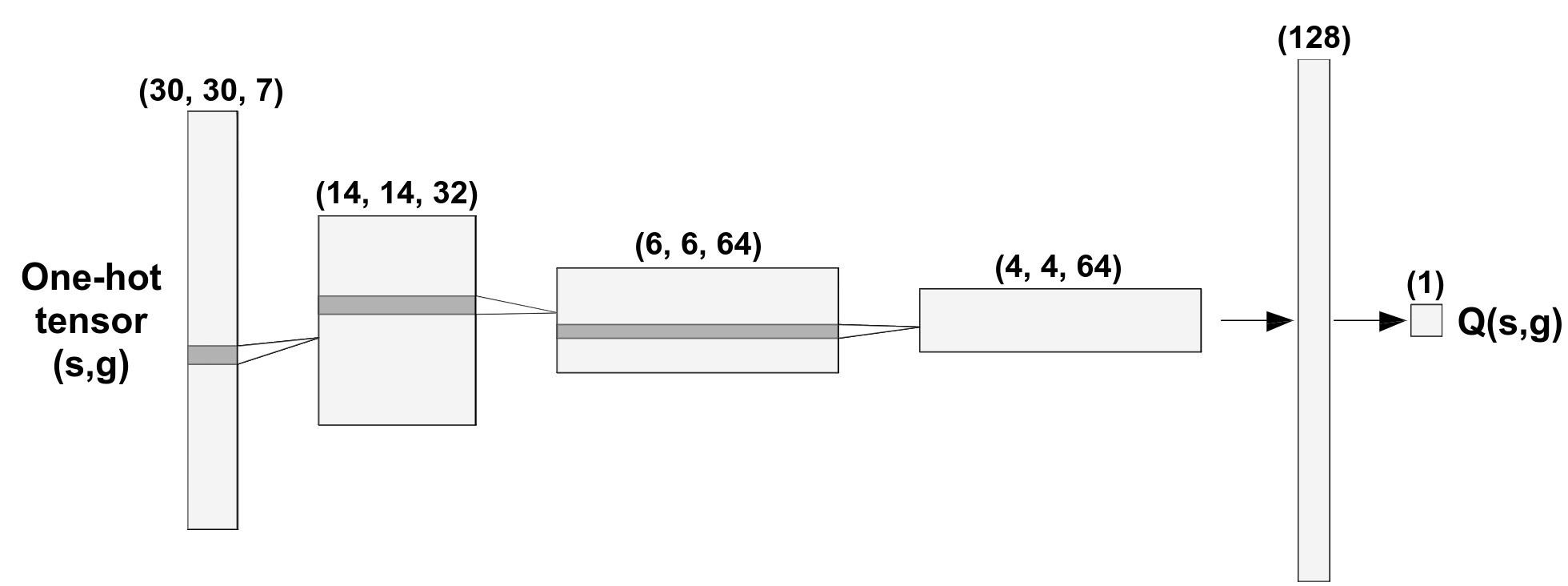}
	\caption{{\textbf{CNN architecture of the Goal Selection Module.} This diagram shows how the size of the one-hot tensor changes as it passes through the layers of the network. The CNN receives an input of size $(30,30,7)$ corresponding to the one-hot tensor of a given $(s,g)$ pair and outputs a single prediction which represents the Q-value $Q(s,g)$.}}
	\label{fig:cnn_architecture}
\end{figure}

{
The loss function $L^g$ is minimized by the DQL algorithm used to train the CNN of the Goal Selection Module. The architecture of this network has been heavily inspired by the one used in the original DQL paper \citep{mnih2013playing}, and is shown in Figure \ref{fig:cnn_architecture}. Initially, the size of the one-hot tensor is $(13,26,7)$, so we increase the lengths of its first two dimensions by adding zeros, i.e., we apply zero-padding, until both have the same size 30. Thus, the CNN receives a square one-hot tensor of size $(30,30,7)$ as input. Then, the CNN applies three convolutional layers. The first layer contains 32 filters and the other two 64 filters each, the same as in \citep{mnih2013playing}. The first two convolutional layers apply kernels of size $4 \times 4$ with a stride of size 2. The third layer uses kernels of size $3 \times 3$ with a stride of size 1. After the convolutional layers, a fully-connected layer with 128 units is applied. Finally, the output layer contains a single unit which outputs the Q-value. We apply batch normalization before each layer of the network except for the output layer. Regarding the DQL algorithm, we have tested different discount factors, and found the best value to be $\gamma = 0.7$. In addition, we have employed several auxiliary techniques to improve the performance of DQL: Fixed Q-targets \citep{mnih2015human} with $\tau=10000$, Double Q-learning \citep{deep2016} and Prioritized Experience Replay \citep{prioritized2016}.
}

{
This CNN is trained in an offline fashion on datasets extracted from 200 training levels, different from the levels used for testing. These datasets are collected by an agent which performs \emph{random exploration} on these levels, i.e., plays the levels by selecting subgoals completely at random. The process is the following. The agent starts at the initial state $s$ of the corresponding level. It selects a random eligible subgoal $\hat{g} \in G_s$ and tries to find a plan $p(s,\hat{g})$ to it. If $\hat{g}$ is attainable, the agent executes the obtained plan until it achieves $\hat{g}$ and arrives at the state $s'$. Then, a new sample of the form $(s, \hat{g}, l_{p(s,g)}, s')$ is created and added to the dataset of the level. In case it is the final sample of the level, i.e., $\hat{g} = g_f$ and there exists a valid plan $p(s,g_f)$, then there is no next state ($s' = \text{Null}$) and a final reward of $r_f$ is added to $l_{p(s,g)}$. If the subgoal $\hat{g}$ is not attainable, no valid plan will be found and a sample of the form $(s, \hat{g}, \lambda, \text{Null})$ is created and added to the dataset. This whole process is repeated until the final goal $g_f$ is selected and achieved, so every sample is part of a trajectory which successfully solves the level. After achieving $\hat{g}$, this process starts again from the initial state of the level. Once 500 unique samples have been gathered, the dataset is saved and this process is repeated for a new level. This algorithm is used to extract the training dataset of each level, for a total of 100000 unique samples, which are used to train the CNN.
}

\section{Experiments and Analysis of Results}
\label{section:experiments_and_analysis_of_results}
{
This section describes the experiments carried out in this work and analyzes the results obtained. The goal of our experimentation is three-fold. Firstly, we assess the quality of the plans obtained with our approach depending on the amount of training data used. Secondly, we compare our model with a state-of-the-art planner. Thirdly, we compare it with the standard Deep Q-Learning algorithm. This way, we are able to evaluate the performance of our model, measured as both plan quality and time requirements, when compared with alternative approaches which try to solve the same problem.
}

\subsection{Evaluating the performance of Deep Q-Planning}
\label{subsection:evaluate_dqp}

{{We have used the Fast-Forward (FF) Planning System \citep{hoffmann2001ff} for our Planner Module because it is a state-of-the-art classical planner, one of the most referenced in the planning literature. Furthermore, it is compatible with PDDL2.1 features such as conditional effects and PDDL functions, which are expressive enough to represent domains such as those of video games.} Among the different search strategies available for FF, we have decided to use best-first search (BFS). Initially, we tried to obtain plans of optimal length, but this proved too computationally expensive for some levels. Thus, we recurred to the evaluation function $f = g + 5*h$ for BFS, where $g$ is the current plan length, and $h$ is an estimation of plan length. This way, the Planner Module is able to obtain plans of near-optimal length in real-time.}

{
{We conducted a first experiment designed to evaluate the performance (measured in plan length) of our planning and acting architecture, as more training data is available. From now on it will be referred to as the Deep Q-Planning (DQP) model.} We trained the DQP model on the datasets extracted from 10, 25, 50 and 100 training levels, randomly selected among the 200 training levels, and finally on the whole training dataset corresponding to all 200 levels. The DQP model was trained for 1.2 million iterations for every dataset size, using the ADAM \citep{adam2015} optimizer with learning rate $\alpha = 1e-05$ and batch size equal to 32. This translated into 3 hours of training time on a machine with a Ryzen 5 3600X CPU and a RTX 2060 GPU when training 5 model instances in parallel. We also used a penalization value of $\lambda = 200$ and a final reward $r_f = -200$ (\emph{see Section \ref{section:goal_selection_learning}}).
}

{For each dataset size, we trained 10 instances of the DQP model, and evaluated each one on 11 test levels, measuring the number of actions needed by the DQP model to solve the test levels. These test levels are different from the ones used for training, in order to measure the generalization ability of the DQP model when applied to levels not seen during training. In addition, they can be grouped into \emph{easy} and \emph{hard} levels. The easy levels correspond to the 5 Boulder Dash levels provided in the GVGAI environment by default. We created 6 additional hard levels to test the DQP model on. These levels were purposefully designed to be as hard to solve by the FF planner as possible, but without increasing the level size. For example, we found that FF had trouble solving levels which contained a large amount of boulders.
}

\begin{figure}[h]
	\centering
	\includegraphics[width=.7\linewidth]{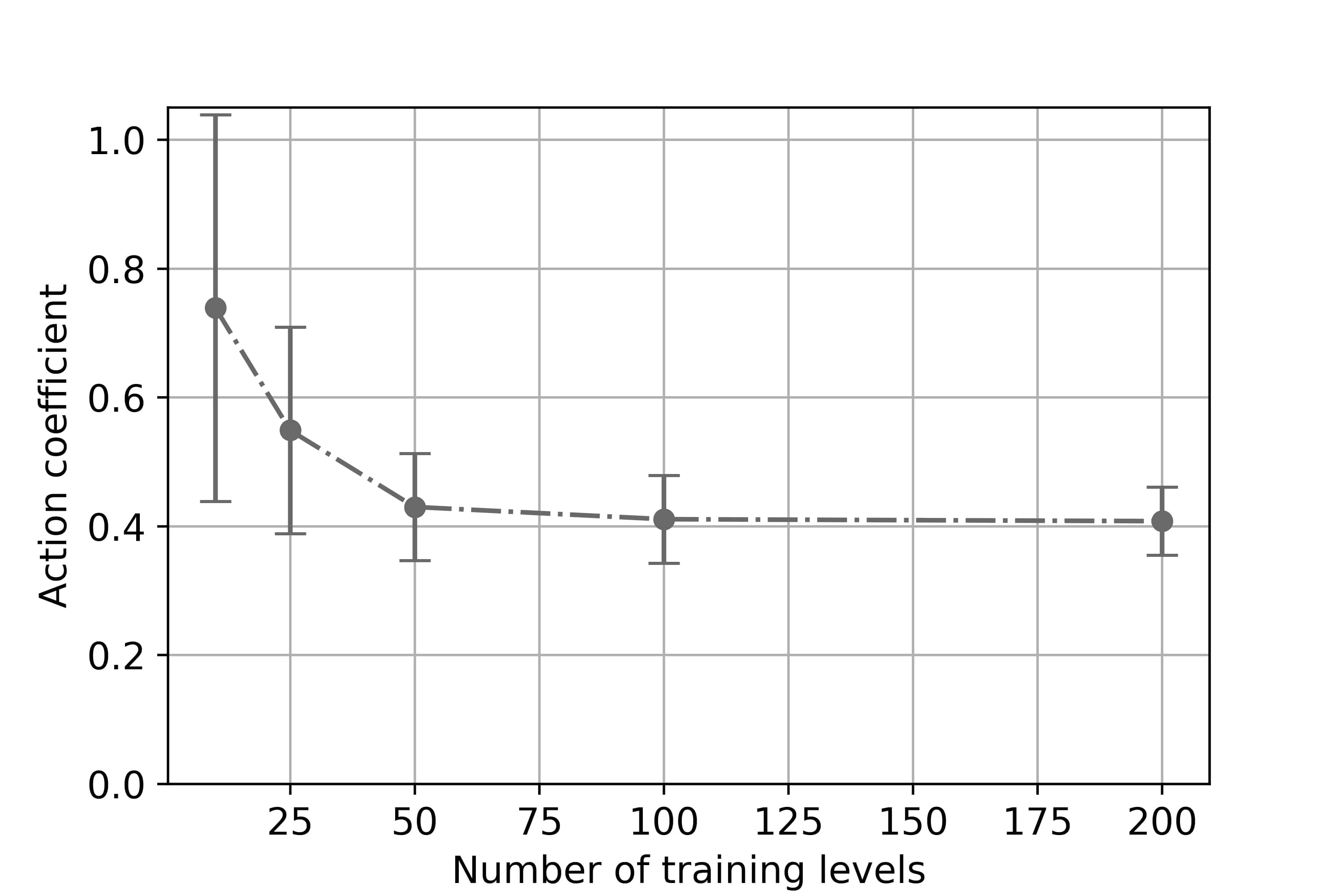}
	\caption{{\textbf{Plan quality of the DQP model for different dataset sizes.} This plot shows the average action coefficient (lower is better) of the DQP model as the number of training levels is increased. Each error bar represents an interval of $\pm1$ standard deviation.}}
	\label{fig:plot_action_coef}
\end{figure}

{
To put the length of the plans obtained by the DQP model into perspective, we tried to compare them with the optimal plan for each test level. However, as mentioned earlier, obtaining the optimal plans proved to be an intractable problem. Thus, we compared the DQP model with a naive, baseline model, which we call the \emph{Random Model} (RM). This model works the same way as the DQP model except for the fact that, instead of selecting the subgoal with lowest Q-value, it selects a $\hat{g} \in G_s$ completely at random. Therefore, for each trained DQP instance, we divided the length of the plans obtained for every test level by the length of those obtained using the Random Model. Then, we calculated the geometric average of this quotient across all the 11 test levels. This way, we obtained for each trained DQP model a metric, called the \emph{action coefficient}, representing the quality of the plans obtained by the model on the test levels. 
{For instance, an action coefficient of 0.6 means that the DQP model uses, on (geometric) average, only 60\% of the actions that the Random Model would use to solve the same 11 test levels.} Figure \ref{fig:plot_action_coef} shows the average action coefficient (across the 10 repetitions) of the DQP model, and its standard deviation as the dataset size (number of training levels) increases.
}

\subsection{Comparing DQP with a state-of-the-art planner}
\label{subsection:dqp_vs_planner}

{
In addition to this first experiment, we conducted a second set of experiments to compare the performance, measured as both plan quality and time requirements, of the DQP model with alternative, state-of-the-art approaches. The results are shown in Table \ref{tab:performance_comparison} and discussed in Section \ref{subsection:discussion}. The DQP model was trained on the whole dataset, corresponding to all 200 training levels. For each test level, we obtained the length of the plan used to solve it and the time needed to obtain it. For the DQP model this time is equal to the sum of the goal selection time and the planning time. The goal selection time is the total time used by the Goal Selection Module to select each subgoal $\hat{g} \in G_s$. The planning time is the total time used by the Planner Module to obtain each plan $p(s,\hat{g})$. We also consider all the time wasted due to goal selection errors (when $\hat{g} \notin G^{\circ}_s$).
}

\begin{table}[h!]
\scriptsize	
\centering
{
\begin{tabularx}{\textwidth}{
>{\centering\arraybackslash}X 
>{\centering\arraybackslash}X 
>{\centering\arraybackslash}X 
>{\centering\arraybackslash}X 
>{\centering\arraybackslash}X 
>{\centering\arraybackslash}X 
>{\centering\arraybackslash}X 
>{\centering\arraybackslash}X 
>{\centering\arraybackslash}X 
>{\centering\arraybackslash}X 
>{\centering\arraybackslash}X 
>{\centering\arraybackslash}X
}

\toprule

\multirow{3}{*}{Models} & \multicolumn{11}{c}{Plan Length (Number of Actions)} \\

& \multicolumn{5}{c}{Easy Levels} & \multicolumn{6}{c}{Hard Levels} \\

\cmidrule(lr){2-6} \cmidrule(lr){7-12}

& 0 & 1 & 2 & 3 & 4 & 5 & 6 & 7 & 8 & 9 & 10 \\

\addlinespace[5pt]

DQP & 85 \newline $\pm7$ & 52 \newline $\pm5$ & 72 \newline $\pm6$ & 84 \newline $\pm12$ & 67 \newline $\pm16$ & 97 \newline $\pm18$ & 137 \newline $\pm15$ & 167 \newline $\pm27$ & 108 \newline $\pm11$ & 89 \newline $\pm12$ & 106 \newline $\pm13$ \\ 

\addlinespace[5pt]

DQL & $-$ & 1165 \newline $\pm0$ & $-$ & $-$ & $-$ & $-$ & $-$ & $-$ & $-$ & $-$ & $-$ \\

\addlinespace[5pt]

RM & 207 \newline $\pm47$ & 188 \newline $\pm59$ & 144 \newline $\pm32$ & 177 \newline $\pm45$ & 190 \newline $\pm46$ & 214 \newline $\pm67$ & 302 \newline $\pm90$ & 456 \newline $\pm101$ & 235 \newline $\pm91$ & 239 \newline $\pm55$ & 262 \newline $\pm74$ \\

\addlinespace[5pt]

BFS & 81 & 67 & 55 & 77 & 43 & $-$ & $-$ & $-$ & $-$ & $-$ & 116 \\

\addlinespace[5pt]

EHC & $-$ & 49 & 44 & 86 & 54 & 117 & $-$ & 140 & 86 & $-$ & $-$ \\

\addlinespace[5pt]

OPT & $-$ & 31 & 42 & $-$ & 38 & $-$ & $-$ & $-$ & $-$ & $-$ & $-$ \\

\bottomrule

\end{tabularx}
}

{
\begin{tabularx}{\textwidth}{
>{\centering\arraybackslash}X 
>{\centering\arraybackslash}X 
>{\centering\arraybackslash}X 
>{\centering\arraybackslash}X 
>{\centering\arraybackslash}X 
>{\centering\arraybackslash}X 
>{\centering\arraybackslash}X 
>{\centering\arraybackslash}X 
>{\centering\arraybackslash}X 
>{\centering\arraybackslash}X 
>{\centering\arraybackslash}X 
>{\centering\arraybackslash}X
}

\toprule

\multirow{3}{*}{Models} & \multicolumn{11}{c}{Time (seconds)} \\

& \multicolumn{5}{c}{Easy Levels} & \multicolumn{6}{c}{Hard Levels} \\

\cmidrule(lr){2-6} \cmidrule(lr){7-12}

& 0 & 1 & 2 & 3 & 4 & 5 & 6 & 7 & 8 & 9 & 10 \\

\addlinespace[5pt]

DQP & 1.39 \newline $\pm0.12$ & 0.54 \newline $\pm0.05$ & 1.41 \newline $\pm0.13$ & 0.57 \newline $\pm0.07$ & 1.35 \newline $\pm0.14$ & 0.74 \newline $\pm0.08$ & 1.49 \newline $\pm0.1$ & 1.09 \newline $\pm0.19$ & 1.58 \newline $\pm0.14$ & 0.6 \newline $\pm0.06$ & 1.43 \newline $\pm0.1$ \\ 

\addlinespace[5pt]

DQL & $-$ & 1.88 \newline $\pm0$ & $-$ & $-$ & $-$ & $-$ & $-$ & $-$ & $-$ & $-$ & $-$ \\

\addlinespace[5pt]

RM & 0.47 \newline $\pm0.1$ & 0.43 \newline $\pm0.12$ & 0.32 \newline $\pm0.06$ & 0.39 \newline $\pm0.1$ & 0.45 \newline $\pm0.1$ & 0.55 \newline $\pm0.12$ & 0.61 \newline $\pm0.16$ & 0.97 \newline $\pm0.2$ & 0.58 \newline $\pm0.2$ & 0.53 \newline $\pm0.11$ & 0.52 \newline $\pm0.12$ \\

\addlinespace[5pt]

BFS & 192.52 & 0.44 & 0.13 & 509.35 & 0.06 & $-$ & $-$ & $-$ & $-$ & $-$ & 95.17 \\

\addlinespace[5pt]

EHC & $-$ & 0.07 & 0.04 & 0.32 & 0.07 & 813.41 & $-$ & 102.21 & 284.06 & $-$ & $-$ \\

\addlinespace[5pt]

OPT & $-$ & 4.75 & 8.62 & $-$ & 305.92 & $-$ & $-$ & $-$ & $-$ & $-$ & $-$ \\

\bottomrule

\end{tabularx}

{
\begin{tabularx}{\textwidth}{
>{\centering\arraybackslash}X 
>{\centering\arraybackslash}X 
>{\centering\arraybackslash}X 
>{\centering\arraybackslash}X 
>{\centering\arraybackslash}X 
>{\centering\arraybackslash}X 
>{\centering\arraybackslash}X 
>{\centering\arraybackslash}X 
>{\centering\arraybackslash}X 
>{\centering\arraybackslash}X 
>{\centering\arraybackslash}X 
>{\centering\arraybackslash}X
}

\toprule

\multirow{3}{*}{Models} & \multicolumn{11}{c}{Success Rate (\%)} \\

& \multicolumn{5}{c}{Easy Levels} & \multicolumn{6}{c}{Hard Levels} \\

\cmidrule(lr){2-6} \cmidrule(lr){7-12}

& 0 & 1 & 2 & 3 & 4 & 5 & 6 & 7 & 8 & 9 & 10 \\

\addlinespace[5pt]

DQL & 0 & 10 & 0 & 0 & 0 & 0 & 0 & 0 & 0 & 0 & 0 \\ 

\bottomrule

\end{tabularx}}

}

\vspace{0.2cm}

\caption{{
\textbf{Comparison of DQP with alternative models}. The table shows mean and std values for plan length (uppermost table)  and time requirements (middle table), in solving easy and hard game levels, by all the models we have compared with. A value of $-$ means a model could not solve the level. In case of BFS, EHC and OPT, this means a 1h timeout was produced. In case of DQL, this means the model could not complete the level under 2000 actions. Since DQL is stochastic we also show values (lower table) for its success rate, by measuring how many times it is able to solve each level out of 10 executions. We have performed 10 repetitions for the DQP, DQL and RM models. Since FF is deterministic, we have performed a single repetition for the BFS, EHC and OPT models.}}
\label{tab:performance_comparison}
\end{table}

{
We have decided to use a classical planner which performs no goal selection whatsoever as one of the approaches to compare the DQP model against. Specifically, we have chosen FF since it is the same planner the DQP model uses. This way, we have applied FF to solve every test level without selecting subgoals, measuring both plan length and planning time. We have tested three different search strategies for FF: best-first search with $f = g + h$ so every obtained plan is optimal in length (OPT model), best-first search with exactly the same evaluation function $f = g + 5*h$ the DQP model uses in order to obtain near-optimal plans (BFS model), and enforced hill-climbing with no optimizations options at all (EHC model). Since FF is deterministic, we have performed a single execution per test level. In addition, we have established a maximum of 1 hour of planning time per level, after which we assume the corresponding model cannot solve the level and a timeout is produced.
}

\subsection{Comparing DQP with standard Deep Q-Learning}
\label{subsection:dqp_vs_dql}

{
We have also compared the DQP model with the standard Deep Q-Learning algorithm, which we will refer to as the DQL model. We have trained the DQL model in an offline fashion, {on datasets extracted from the same 200 training levels as DQP, and evaluated it on all 11 test levels, measuring both plan lengths and action selection times.} Just as with DQP, we have repeated each execution 10 times. We have also established a maximum of 2000 actions per each test level. If the DQL model cannot complete a level under 2000 actions, we assume it cannot solve that level.
}

{
In order to extract the dataset for each training level, we have not used the $\epsilon$-greedy exploration-exploitation strategy commonly employed in RL. {This is because, unlike most RL setups, we separate training (exploration) and test (exploitation) in two distinct phases, using different levels for each phase.} Thus, we have designed an algorithm inspired by $\epsilon$-greedy but adapted to our problem. The agent starts at the initial state of the level, selects a random subgoal $\hat{g} \in G_s$ (gem), obtains the plan $p(s,\hat{g})$ and executes it. Once achieved $\hat{g}$, the agent executes a number $n$ of random actions, where $n$ has been uniformly sampled from 1 to 10. After executing the random actions, it obtains and executes the plan to another random subgoal. These two phases (plan execution and random walk) interleave until the agent achieves $g_f$ and solves the level, at which point this process starts again from the initial state.  {Each time the agent executes an action, a sample is collected and once 5000 unique samples have been gathered the dataset is saved.} This algorithm is used to extract the training dataset of the DQL model for each of the 200 levels, for a total of one million unique samples, ten times more samples than for the DQP model.  {The plan execution phase guarantees each trajectory always solves the level (the agent ultimately achieves $g_f$), whereas the random walk phase helps the agent explore all the state space.} This way, we are able to obtain samples with both good diversity and quality.
}

{These samples are of the form $(s,a,r,s')$ and correspond to the DMDP $M$, which describes the game from a standard RL perspective (\emph{see Section \ref{subsection:math_formulation}}).
Each sample is interpreted as follows: the agent is in a state of the game $s$, it then executes an action $a$, among the set of possible actions $A = \{UP, DOWN, LEFT, RIGHT, USE\}$, arriving at the next state $s'$, and obtaining the immediate reward $r$. The agent receives a reward $r = 5$ after completing the level (in this case $s' = Null$), and obtains a reward $r = -1$ otherwise (the
agent is penalized for using too many actions to solve the level). It’s important to note that, unlike the DQP model, planning is only used to help collect samples. The DQL model itself does not learn to select subgoals, and does not perform any time of planning when solving the test levels.
}

{
The architecture and hyperparameters are the same for both the DQP and DQL models, except for the following differences:}
\begin{itemize}
    
    \item  {The DQL model iterates over every eligible action $a \in A$ and predicts the Q-value $Q(s,a)$ associated with each one. Then, the action with the highest Q-value is selected. The CNN of the DQL model receives as inputs both the current state $s$ and a possible action $a$, and returns the predicted Q-value $Q(s,a)$ (\emph{see Equation \ref{equation:deep-q-learning}}). 
    {The state $s$ is encoded as a one-hot tensor, whereas the action $a$ is encoded as a separate one-hot vector, and does not form part of the one-hot tensor.}
    }
    
    \item { The DQL model needs to know the orientation of the agent (\emph{NORTH}, \emph{EAST}, \emph{SOUTH} or \emph{WEST}), which is also encoded in a one-hot vector. These two one-hot vectors, corresponding to the action and the orientation, are concatenated to the flattened output of the third convolutional layer, and together are given as input to the first fully-connected layer. Since the DQL model selects actions instead of subgoals like DQP, game trajectories are effectively longer and, thus, we need to apply a smaller discount to rewards (use a bigger value for $\gamma$). We have fixed this new discount factor to $\gamma=0.99$, the value used in \citep{mnih2013playing}. 
    {In addition, we experimented with different learning rates and number of training iterations (up to 10 million). We found that $\alpha=5e-06$ was the optimal learning rate, and 6.5 million the best number of training iterations.}
}

\item { Finally, we have implemented a mechanism for detecting and avoiding loops. A loop occurs when the DQL model arrives at a state $s'$ which has already been visited. {Since the agent is deterministic, it will execute the exact same sequence of actions from that point onward, and will be forever trapped in a loop.} In order to avoid this, our algorithm saves a record of the states visited by the agent and how many times they have been visited. If the agent arrives at an unvisited state $s'$, the action with the highest Q-value is selected. If $s'$ has already been visited, then the agent selects the action with the highest Q-value which has not been tried yet. If every action has been executed, then the agent simply takes a random action. This way, the agent can escape loops. However, due to this loop detection mechanism, the behaviour of the DQL model becomes stochastic. For this reason, apart from action selection time and plan lengths, we have also measured its success rate for each test level, i.e., how many times it is able to solve each level (using fewer than 2000 actions) out of the 10 executions.
}
\end{itemize}
\subsection{Discussion}
\label{subsection:discussion}

{
As it is shown in Figure \ref{fig:plot_action_coef}, the action coefficient of the DQP model decreases as the number of training levels increases. This means that the model performs better, i.e., obtains shorter plans, as it is trained on more data, as was to be expected. The performance of the model improves rapidly from 10 to 50 training levels. {It slows down from 50 to 100 levels, and there is only a slight improvement when using 200 training levels instead of only 100.} Thus, the DQP model is able to obtain a close-to-optimal performance when trained on 50000 samples split across 100 levels, reaching an action coefficient of 0.41. This means that, on average, it obtains plans with only 41\% of the number of actions used by a model which selects subgoals at random. 
}

{
As it will be later argued, these are quite remarkable results, since they show that the DQP model is able to properly generalize what has learned in the training levels to the previously unseen test levels, while needing only a fraction of the training data which standard RL algorithms use. Finally, we want to mention that, as more training data is used, the standard deviation of the action coefficient also reduces. For a dataset of 100 training levels, the standard deviation is equal to 0.068 and, for 200 levels, is equal to 0.053. This means that, when trained on enough data, the DQP model is able to obtain consistent, stable results across all executions.
}

{
Table \ref{tab:performance_comparison} compares the results obtained by the different models, in terms of plan lengths, time requirements and success rate (this last metric only used for the DQL model). Regarding time requirements, we can observe how the DQP model is able to solve every test level under two seconds of total time, which is equal to the sum of goal selection and planning times. Besides, there is no significant difference between the times of easy and hard levels. The Random Model also obtains similar times across all levels, although these times are smaller than those of the DQP model. {This happens because the RM model does not spend time selecting subgoals, but rather selects them at random.}
}

{
The behaviour of the models based on classical planning (BFS, EHC and OPT) is very different. Planning times for these models vary greatly depending on the particular level, not being able to solve many of them before the 1 hour timeout. The BFS model, which uses the exact same search strategy as the Planner Module of the DQP model, can only solve three easy levels in reasonable time, approximately needs 1.5, 3 and 8 minutes to solve levels 10, 0 and 3, respectively, and cannot solve five out of the six hard levels. The EHC model can only solve four easy levels in a reasonable period of time, can find plans for three hard levels but while spending much more planning time and, finally, cannot solve one easy level and three hard levels. Lastly, the OPT model can only find the optimal plans for three easy levels, needing more than five minutes for one of them.
}

{
If we now compare the plan lengths of the different models, we can observe how the DQL model is only able to solve one level (level 1) while using fewer than 2000 actions. In addition, it has a success rate of 10\% for that level, meaning that it is only able to do it in one out of the ten executions. As these results surprised us, we performed additional experimentation for the DQL model in order to validate the model. We designed a simple level consisting of a small confined area with only 23 gems, the agent and the exit. We collected 30000 samples from this level and trained the DQL model only on the dataset extracted from this level. The DQL model was able to consistently solve this simple level, although not in an optimal way. Therefore, the bad performance of the DQL model shows that the standard Deep Q-Learning algorithm needs to be trained on more than just one million samples (and possibly also for a bigger number of training iterations) to be able to successfully solve the Boulder Dash game, especially when trained and tested on different sets of levels.
}

{
Unlike DQL, the DQP model is able to solve each test level and obtain plans of good quality. It {reaches} an action coefficient of 0.40 meaning that, by learning to select subgoals with Deep Q-Learning, we are able to {obtain} plans on average 60\% shorter than those obtained by selecting subgoals at random. If we 
divide the length of the plans obtained by the DQP model by those obtained using the BFS model, not considering those levels BFS cannot solve, and calculate the geometric average (as we did to obtain the action coefficient), we obtain a value of 1.09. If we repeat the same calculation for the EHC model, we now obtain a value of 1.15. These values mean that, on average, the DQP model obtains plans with 9\% and 15\% more actions than the BFS and EHC models, respectively.
}

{
To sum up, the results of our experiments show how the DQP model performs better than standard Deep Q-Learning and classical planning, when both plan quality and time requirements are considered at the same time. On the one hand, we are able to drastically increase the performance of Deep Q-Learning by applying it to select subgoals {instead of actions, and using} a planner to {achieve} the selected subgoals. Results show how the DQP model greatly outperforms DQL while being trained on a dataset ten times smaller. 
{Thus, we can conclude our DQP approach is at least one order of magnitude more sample-efficient than standard DQL. Moreover, it also generalizes better when applied to levels not seen during training.} Our hypothesis is that this happens because, as mentioned in Section \ref{subsection:math_formulation}, by formulating this problem using our goal selection approach (described by $M^g$) instead of the standard RL formulation (given by $M$), the state space is reduced ($S^g \subset S$) and the learning problem is simplified.
}

{
On the other hand, the comparison among DQP and the models based on classical planning has shown how our 
goal selection approach is able to substantially reduce the time requirements of classical planning for most levels. In addition, it achieves consistent,  {stable times across all 11 test levels, whereas FF exhibits drastic variation in time performance, depending on the test level and the search strategy applied.} This comes at the expense of a small decrease in plan quality, as our approach obtains plans with 9\% more actions than the BFS model, which applies the exact same search strategy as the Planner Module of the DQP model.
}

\section{Related Work}
\label{section:related_work}

The use of Neural Networks (NN) in Automated Planning has been a topic of great interest in recent years. Some works have applied Deep Q-Learning to solve planning and scheduling problems as a substitute for online search algorithms. \citep{shen2017deep} uses Deep Q-Learning to solve the \emph{ship stowage planning problem}, i.e., in which slot to place a set of containers so that the slot scheme satisfies a series {of constraints, and optimizes several} objective functions at the same time. \citep{mukadam2017tactical} also employs Deep Q-Learning, but this time to solve the \emph{lane changing problem}. In this problem, autonomous vehicles must automatically change lanes in order to avoid the {traffic, and get} to the exit as quickly as possible. Here, Deep Q-Learning is only used to learn the long-term strategy, while relying on a low-level module to change between adjacent lanes without collisions. In our work, we also employ Deep Q-Learning but, instead of using it as a substitute for classical planning, we integrate it along with planning into our planning and acting architecture.

There are other works which use neural networks to solve planning problems but, instead of relying on RL techniques such as Deep Q-Learning, train a NN so that it learns to perform an \emph{explicit planning process}. \citep{toyer2018action} proposes a novel NN architecture known as \emph{Action Schema Networks} (ASNet) which, as they explain in their work, \emph{are specialised to the structure of planning problems much as Convolutional Neural Networks (CNN) are specialised to the structure of images}. \citep{tamar2016value} uses a CNN that performs the computations of the value-iteration (VI) planning algorithm \citep{bellman1958dynamic,bertsekas2017dynamic}, thus making the planning process differentiable. This way, both works use NN architectures which \emph{learn to plan}. 

These NNs are trained on a set of training problems and evaluated on different problems of the same planning domain, showing better generalization abilities than most RL algorithms. \citep{tamar2016value} argues that this happens because, in order to generalize well, NNs need to learn an \emph{explicit planning process},  which most RL techniques do not. Although our architecture does not learn to plan, it does incorporate an off-the-shelf planner which performs explicit planning. We believe this is why our architecture shows good generalization abilities.

Neural networks have also been applied to other aspects of planning. For instance, \citep{dittadi2018learning} trains a NN that learns a planning domain just from visual observations, assuming that actions have local preconditions and effects. The learned domain is generalizable across different problems of the same domain and, thus, can be used by a planner to solve these problems.

{
There exist several techniques which facilitate the application of Automated Planning in real-time scenarios, such as Goal Reasoning \citep{aha2018goal}, Anytime Planning \citep{richter2010lama}, Hierarchical Planning (e.g., HTN \citep{georgievski2015htn}) and domain-specific heuristics learned using ML \citep{yoon2008learning}. \citep{guzman2012pelea} presents PELEA, a domain-independent, online execution architecture which performs planning at two different levels, \emph{high} and \emph{low}, and is able to learn domain models, low-level policies and planning heuristics. \citep{mcgann2008deliberative} proposes T-REX, an online execution system used to control autonomous underwater vehicles. This system partitions deliberation across a set of concurrent \emph{reactors}. Each reactor solves a different part of the planning problem and cooperates with the others, interchanging goals and state observations.
}

{
Some works incorporate Goal Selection into planning and acting architectures. \citep{jaidee2012learning} proposes a Goal Reasoning architecture which combines Case-Based Reasoning with Q-Learning. In our work, we have focused on learning to select subgoals, using Deep Q-Learning instead of traditional Q-Learning, in order to give our architecture the ability to generalize to new states. \citep{bonanno2016selecting} makes use of a CNN which learns to select subgoals from images. Unlike our work, the CNN is trained by a hard-coded expert procedure in a supervised fashion, and the set of eligible subgoals is always the same, regardless of the state of the game.
}

Finally, it is worth to mention previous disruptive work on Deep RL \citep{mnih2015human} that addresses how to learn  models to control the behavior of reactive agents in ATARI games. As opposite to this work, we are interested in addressing how deliberative behaviour (as planning is) can be improved by mainstream techniques in Machine Learning. This is one of the main reasons we chose the GVGAI video game framework, since it provides an important repertory of video games where deliberative behaviour is mandatory to achieve a high-level performance.

\section{Conclusions and Future Work}
\label{section:conclusions_and_future_work}

{
In this work we have proposed a planning and acting architecture which combines Reinforcement Learning with Automated Planning. It learns to select subgoals using Deep Q-Learning, which are then achieved with the help of a classical, PDDL-based planner. It has been trained on a deterministic version of the game known as Boulder Dash, using different levels for training and testing in order to measure its generalization abilities. We have conducted an experiment to measure how the quality (length) of the plans obtained with our approach improves as more training data is made available. Additionally, we have compared the performance of our model, in terms of both plan quality and time requirements, with that of standard Deep Q-Learning and classical planning methods.
}

{
The obtained results show our DQP model is able to find plans of good quality while meeting real-time requirements. Thanks to goal selection, it is able to exploit the synergy between AP and RL and obtain better results than any of these techniques on their own.
On the one hand, we adapt DMDPs to our goal selection and planning approach, training the Deep Q-Learning algorithm to select subgoals instead of actions. 
This way, we are able to improve sample-efficiency and generalization across levels, obtaining plans of better quality than standard Deep Q-Learning when trained on ten times more data.
On the other hand, our DQP model substantially reduces the time requirements of Automated Planning, at the expense of obtaining plans with 9\% more actions on average.
By using goal selection, we are able to solve every game level under 2 seconds of total time. However, when no goal selection is performed, the same planner our architecture uses can only solve 3 out of 11 levels in reasonable time.
}

{
In future work we want to extend this experimentation and test our DQP model on additional games, including non-deterministic games such as the original version of Boulder Dash. In order to do this, our approach must be able to manage uncertainty. We plan to use Deep Q-Learning not only to predict plan length, but also the uncertainty associated with a given subgoal. If a subgoal has a large uncertainty value, this means that a plan from the current state to this subgoal is likely to fail, e.g., due to an obstacle moving and blocking the path of the agent. We intend to use this uncertainty value for selecting subgoals as well as monitoring plan execution, so a new subgoal can be selected if an unexpected situation arises.
}

\section{Acknowledgements}
\label{section:acknowledgements}

We thank Vladislav Nikolov-Vasilev for the implementation of the PDDL Parser {and Ignacio Vellido-Expósito for his aid in performing the experiments.}

Funding: This work was supported by the Spanish MINECO R\&D Project [RTI2018-098460-B-I00]; and EU FEDER Funds.

\bibliographystyle{apalike}
\bibliography{References}

\end{document}